\documentclass{article}

 \usepackage[sglblindworkshop, final]{neurips_2025}
\workshoptitle{First Workshop on the Foundations of Reasoning in Language Models (FoRLM)}
\usepackage[utf8]{inputenc} % allow utf-8 input
\usepackage[T1]{fontenc}    % use 8-bit T1 fonts
\usepackage{hyperref}       % hyperlinks
\usepackage{url}            % simple URL typesetting
\usepackage{booktabs}       % professional-quality tables
\usepackage{amsfonts}       % blackboard math symbols
\usepackage{nicefrac}       % compact symbols for 1/2, etc.
\usepackage{microtype}      % microtypography
\usepackage{xcolor}         % colors

\title{Formatting Instructions For NeurIPS 2025}

\usepackage{microtype}
\usepackage{hyperref}
\usepackage{url}
\usepackage{booktabs}

\usepackage{caption}
\usepackage{subcaption}
\usepackage{lineno}

\usepackage{url}

\usepackage[utf8]{inputenc} % allow utf-8 input
\usepackage[T1]{fontenc}    % use 8-bit T1 fonts
\usepackage{url}            % simple URL typesetting
\usepackage{booktabs}       % professional-quality tables
\usepackage{amsfonts}       % blackboard math symbols
\usepackage{nicefrac}       % compact symbols for 1/2, etc.
\usepackage{microtype}      % microtypography
\usepackage{xcolor}         % colors
\usepackage{graphicx}
\usepackage{amsmath}
\usepackage{wrapfig}
\usepackage{listings}
\usepackage{svg}
\usepackage{algorithmic}
\usepackage{algorithm}
\usepackage{amsmath}
\usepackage{amssymb}
\usepackage{mathtools}
\usepackage{amsthm}
\usepackage{graphics}
\usepackage{xcolor,pifont}
\usepackage[capitalize,noabbrev]{cleveref}

\definecolor{codegreen}{rgb}{0,0.6,0}
\definecolor{codegray}{rgb}{0.5,0.5,0.5}
\definecolor{codepink}{RGB}{252, 142, 172}
\definecolor{codepurple}{rgb}{0.58,0,0.82}
\definecolor{backcolour}{RGB}{245,245,245}
\lstdefinestyle{mystyle}{
    backgroundcolor=\color{backcolour},   
    commentstyle=\color{magenta},
    keywordstyle=\color{blue},
    numberstyle=\tiny\color{codegray},
    stringstyle=\color{codepurple},
    basicstyle=\fontfamily{\ttdefault}\footnotesize,
    breakatwhitespace=false,         
    breaklines=true,                 
    keepspaces=true,    
    frame=single,
    numbersep=5pt,                  
    showspaces=false,                
    showstringspaces=false,
    showtabs=false,                  
    tabsize=2,
    classoffset=1, % starting new class
    keywordstyle=\color{violet},
    classoffset=0,
}
\lstdefinelanguage{JavaScript}{
  keywords={typeof, new, true, false, catch, function, return, null, catch, switch, var, if, in, while, do, else, case, break},
  keywordstyle=\color{black},
  ndkeywords={class, export, boolean, throw, implements, import, this},
  ndkeywordstyle=\color{darkgray}\bfseries,
  identifierstyle=\color{black},
  sensitive=false,
  comment=[l]{//},
  morecomment=[s]{/*}{*/},
  commentstyle=\color{purple}\ttfamily
}
\definecolor{darkblue}{rgb}{0, 0, 0.5}
\hypersetup{colorlinks=true, citecolor=darkblue, linkcolor=darkblue, urlcolor=darkblue}

\title{Exploration with Foundation Models: Capabilities, Limitations, and Hybrid Approaches}

\author{
  Remo Sasso \quad Michelangelo Conserva \quad Dominik Jeurissen \quad Paulo Rauber \\
  School of Electronic Engineering and Computer Science\\
  Queen Mary University of London, United Kingdom \\
  \texttt{\{r.sasso, m.conserva, d.jeurissen, p.rauber\}@qmul.ac.uk}
}

\begin{document}

\maketitle
\begin{abstract}
Exploration in reinforcement learning (RL) remains challenging, particularly in sparse-reward settings. While foundation models possess strong semantic priors, their capabilities as zero-shot exploration agents in classic RL benchmarks are not well understood.We benchmark LLMs and VLMs on multi-armed bandits, Gridworlds, and sparse-reward Atari to test zero-shot exploration.
 Our investigation reveals a key limitation: while VLMs can infer high-level objectives from visual input, they consistently fail at precise low-level control—the “knowing–doing gap”. To analyze a potential bridge for this gap, we investigate a simple on-policy hybrid framework in a controlled, best-case scenario. Our results in this idealized setting show that VLM guidance can significantly improve early-stage sample efficiency, providing a clear analysis of the potential and constraints of using foundation models to guide exploration rather than for end-to-end control.
\end{abstract}

\section{Introduction}
\label{sec:introduction}
Reinforcement learning (RL) provides a framework for sequential decision-making, where an agent interacts with an environment to maximize cumulative rewards \citep{sutton2018reinforcement}. A fundamental challenge in RL is exploration—the need to efficiently discover high-value states rather than prematurely exploiting suboptimal strategies. In sparse-reward settings, where rewards are infrequent, traditional exploration heuristics such as random actions or uncertainty-based methods can be highly inefficient, often requiring millions of interactions to uncover meaningful solutions \citep{bellemare2016unifying}. While advanced model-based methods like posterior sampling have been developed to improve this sample efficiency \citep{sasso2023posterior}, our work explores an alternative direction by leveraging the semantic priors of large foundation models.

Large language models (LLMs) and vision-language models (VLMs) have recently demonstrated strong reasoning, semantic understanding, and in-context learning capabilities \citep{brown2020language, team2023gemini, achiam2023gpt, touvron2023llama, jiang2024mixtral, team2024gemma, deepseekai2025deepseekv3technicalreport}. Unlike RL agents, which rely on trial-and-error learning, LLMs can infer objectives, recognize patterns, and generate structured action sequences with minimal experience \citep{llm_survey, du2023guiding, voyager}. This raises an important question:
\textit{How do LLMs and VLMs perform in traditional hard-exploration settings, and could they be leveraged to improve performance?}

Recent studies have established that LLMs can perform in-context exploration in multi-armed bandits (MABs), though performance is sensitive to complex prompting and history summarization \citep{krishnamurthy2024largelanguagemodelsexplore}. Our work complements these findings by analyzing the impact of simple, general-purpose instruction phrasing (implicit vs. explicit). However, as bandits lack state transitions and long-term planning, a broader evaluation is necessary to assess their potential for general RL. To address this gap, we conduct a systematic benchmark of foundation models as zero-shot agents across a progression of classic exploration environments: from Bernoulli MABs, to spatial reasoning in Gridworlds, and finally to high-dimensional, sparse-reward Atari games.

This paper makes the following contributions:
\begin{itemize}
    \item \textbf{A Systematic Benchmark of Foundation Models in RL Exploration.} We evaluate a range of LLMs and VLMs on a progression of classic exploration tasks to provide a clear empirical snapshot of their zero-shot capabilities.
    \item \textbf{A Characterization of VLM Failure Modes.} Through a detailed qualitative analysis of VLM agents in hard-exploration Atari games, we identify and document a persistent "knowing-doing gap," encompassing challenges in motor control, semantic grounding, and contextual reasoning.
    \item \textbf{An Upper-Bound Analysis of a Hybrid Approach.} We provide a quantitative analysis of a simple hybrid framework in a carefully selected, near-ideal environment. This controlled study serves to establish a potential upper bound on the sample efficiency gains achievable with this approach, rather than to propose it as a general-purpose solution.
\end{itemize}

To investigate these contributions, we conduct a systematic evaluation across a hierarchy of increasingly complex environments. We begin with Multi-Armed Bandits to isolate the exploration-exploitation trade-off, move to Gridworlds to study structured, memory-based exploration, and finally, extend our analysis to hard-exploration Atari games using Vision-Language Models to assess their capabilities with high-dimensional visual input.

The remainder of this paper is organized as follows. Section \ref{sec:related_works} reviews related literature. Section \ref{sec:analysis} presents our benchmark of text-based agents in bandits and Gridworlds. Section \ref{sec:vlmatari} provides our benchmark of vision-language models in Atari, including a detailed qualitative analysis of their failure modes. Section \ref{sec:hybrid_upper_bound} investigates our hybrid framework as a proof-of-concept. Finally, Section \ref{sec:conclusion} summarizes our findings.

\section{Related Work} 
\label{sec:related_works}

Recent research has explored the integration of foundation models into reinforcement learning from multiple perspectives. We position our work in relation to two primary research threads: the evaluation of foundation models as zero-shot decision-makers and the development of hybrid frameworks that combine them with traditional RL agents.

\subsection{Foundation Models as Zero-Shot Decision-Makers}

A growing body of work evaluates the innate capabilities of foundation models as autonomous agents. In the classic exploration-exploitation testbed of multi-armed bandits, studies by \citet{krishnamurthy2024largelanguagemodelsexplore} and \citet{wu2023smartplay} have shown that LLMs can execute exploration strategies, though performance is highly sensitive to detailed prompting. The trend towards more comprehensive evaluation is exemplified by benchmarks like AgentBench \citep{liu2024agentbench}, which systematically assessed LLM agents across eight distinct environments and found that even capable models exhibit aimless behavior in long-horizon tasks.

In more complex visual domains, recent benchmarks have sought to understand VLM capabilities. The Atari-GPT benchmark \citep{waytowich2024atarigptbenchmarkingmultimodallarge} evaluated VLMs on dense-reward Atari games, finding that they struggle with the precise temporal reasoning required for fine-grained control. Concurrently, the BALROG benchmark \citep{paglieri2025balrog} identified a significant \textit{knowing-doing gap} in complex procedural games like NetHack, where models can often describe the optimal strategy but fail to execute the necessary low-level actions. To further isolate this issue, the TextAtari benchmark \citep{li2025textatari} circumvents pixel-based control by converting Atari states into rich textual descriptions. This work demonstrates that when freed from visual-grounding challenges, text-based LLMs can reason and plan much more effectively over long horizons. Collectively, these studies reinforce the idea that the primary bottleneck for current VLMs is not a lack of high-level understanding, but a failure in low-level, pixel-to-action control. Our work directly builds on these insights by conducting a targeted analysis on a suite of classic \textit{sparse-reward}, hard-exploration Atari games, specifically probing VLM reasoning where extrinsic signals are rare.

\subsection{Hybrid Frameworks for RL and Foundation Models}

To bridge the knowing-doing gap, a prominent research direction integrates foundation models with traditional RL algorithms. These hybrid frameworks leverage the strengths of both paradigms. One popular approach is to use the foundation model as an auxiliary component for shaping intrinsic rewards, as seen in Motif \citep{klissarov2024motif}, where an LLM's judgment of "interestingness" guides an RL agent to SOTA performance in NetHack. 

Another powerful approach uses FMs to guide the exploration process itself. For example, Intelligent Go-Explore (IGE) \citep{lu2025intelligent} replaces the hand-crafted heuristics of the Go-Explore algorithm with a GPT-4 model that identifies promising states to return to, solving hard-exploration games where prior LLM agents failed. Other sophisticated frameworks like ACE \citep{wan2025ace} integrate the LLM as an offline planner and critic in an actor-critic loop, enabling improved performance on industrial-scale control problems. 

In contrast to these more complex integrations, our work investigates a simpler, direct-control intervention. We propose an on-policy hybrid framework where the VLM acts as a temporary, exploratory guide for a standard PPO agent. The goal is not to engineer a reward function or build a co-evolutionary system, but rather to use the VLM's zero-shot semantic understanding to steer the agent to promising regions of the state space during the earliest stages of training. This allows us to directly measure the impact of VLM guidance on sample efficiency without introducing complex hierarchical structures.

\section{LLM Exploration Study} \label{sec:analysis}

In this section, we evaluate the performance of LLMs in two structured exploration settings: multi-armed bandits (MABs) and Gridworlds. We investigate whether LLMs can infer the need for exploration through prompt phrasing in MABs (Section \ref{sec:bandits}) and how they adapt to increasing prompt complexity in Gridworlds (Section \ref{sec:gridworld_experiments}).

\begin{figure}
\begin{lstlisting}[breaklines=true, language=JavaScript, caption=The prompt versions used for the bandit settings., label={list:bandit_prompt}]
// Implicit (v1)
Your goal is to maximize the total reward by pulling the arm with the highest probability of success.
// Explicit (v2)
Your goal is to maximize the total reward by finding out which arm has the highest probability of success.
\end{lstlisting}
\end{figure}

\subsection{Multi-Armed Bandits} \label{sec:bandits}
Multi-Armed Bandits (MABs) provide a simplified framework for studying exploration strategies in reinforcement learning (RL). Since bandits do not involve state transitions, they isolate the exploration-exploitation trade-off, making them a useful testbed for evaluating how LLMs reason about exploration. In our experiments, we use Bernoulli bandits, where each arm provides rewards sampled from a Bernoulli distribution. A formal definition of MABs, including the Bernoulli bandits used in our experiments, is provided in Appendix \ref{app:multiarmedbandits}. 

Prior studies have analyzed LLM exploration in bandit settings but primarily focused on complex reasoning techniques, such as external summarization and chain-of-thought prompting, rather than studying the effect of simple prompt phrasing and its impact on inference capabilities \citep{krishnamurthy2024largelanguagemodelsexplore}. To address this gap, we investigate whether LLMs can infer the need for exploration from how instructions are phrased.

\paragraph{Prompts.} We compare two general-purpose prompting strategies, \textit{implicit prompting (v1)} and \textit{explicit prompting (v2)}, each with a different level of specificity (Listing \ref{list:bandit_prompt}). Implicit prompting (v1) provides a broad objective, requiring LLMs to infer exploration needs, whereas explicit prompting (v2) directly instructs exploration. By contrasting these two conditions, we analyze whether LLMs naturally adopt exploration strategies or require explicit guidance to behave optimally.

\begin{figure}[t]
    \centering
    \includegraphics[width=.95\linewidth]{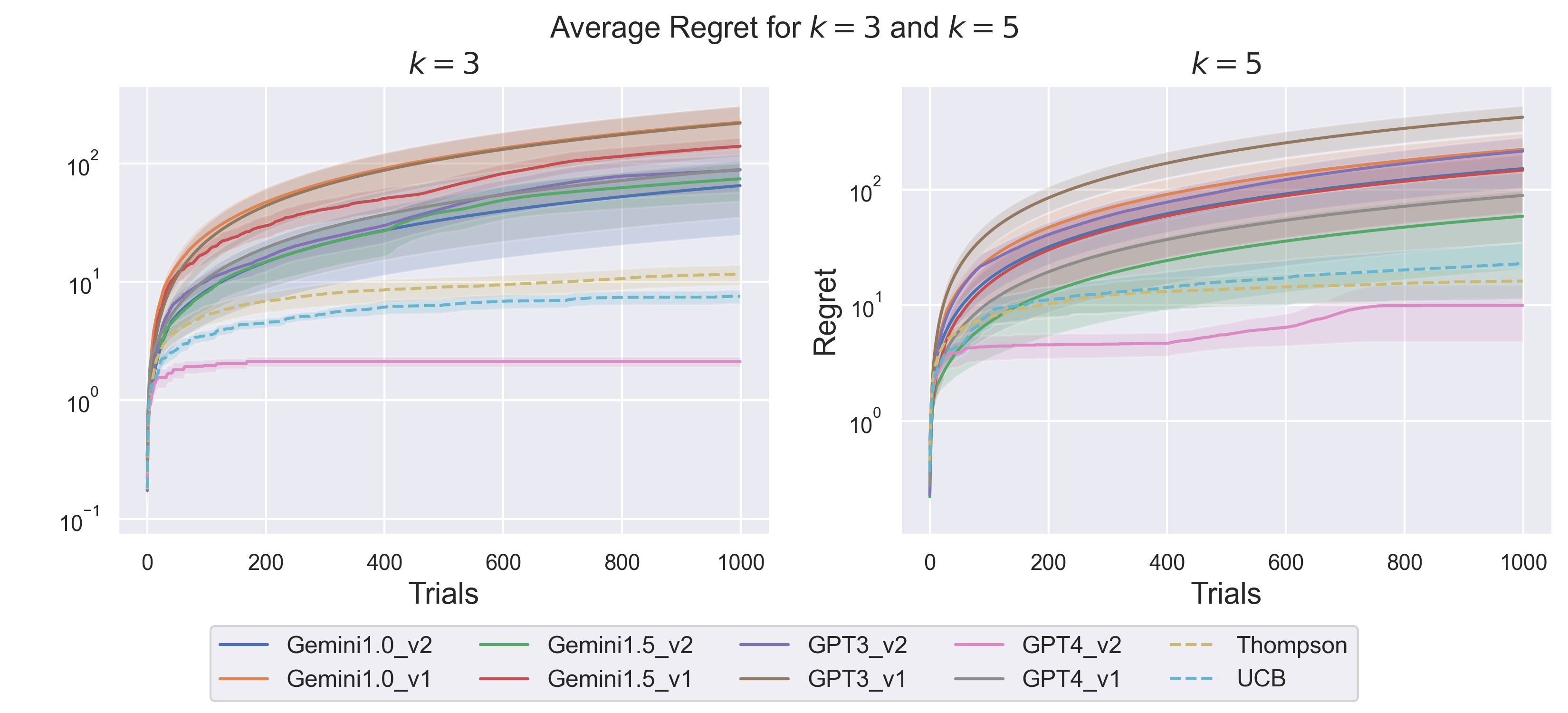}
    \caption{Averaged regret of the bandit experiments with $k=3$ and $k=5$ for various LLMs, Thompson Sampling, and UCB.}
    \label{fig:bandit_results}
\end{figure}

\paragraph{Algorithms.} We evaluate four LLMs—GPT-3.5, GPT-4, Gemini 1.0, and Gemini 1.5—spanning different architectures and parameter scales \citep{achiam2023gpt, team2023gemini}. These models were selected based on API availability and performance comparisons in prior work \citep{chiang2024chatbot}. As baselines, we use two classical exploration algorithms: Thompson Sampling \citep{thompson1933likelihood} and Upper Confidence Bound (UCB) \citep{auer2002finite}, which provide strong theoretical guarantees for exploration efficiency. Further details on these algorithms are provided in Appendix \ref{app:multiarmedbandits}.   

\begin{wrapfigure}[24]{r}{0.4\linewidth}
    \centering
        \includegraphics[width=\linewidth]{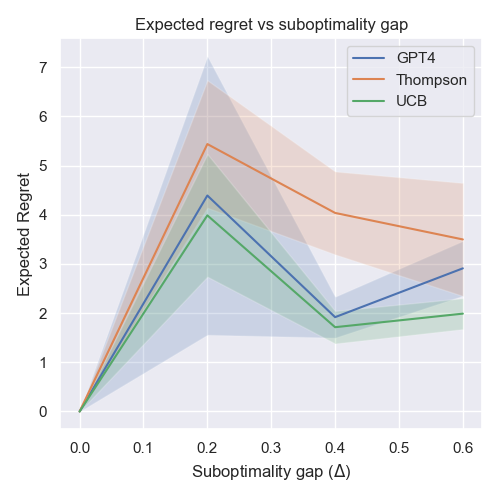}
    \caption{Suboptimality gap experiments for $2$-armed bandit settings, comparing GPT-4, UCB, and Thompson Sampling.}
    \label{fig:suboptimality}

\end{wrapfigure}

\paragraph{Evaluation.} We first conduct a general evaluation of LLM performance in multi-armed bandits by analyzing their behavior in three-armed and five-armed bandit settings. In this setup, the success probabilities of each Bernoulli arm are sampled from $U(0,1)$, rather than using fixed reward structures. This ensures greater variability across runs, allowing us to test whether LLMs can explore effectively in unstructured reward distributions. To assess performance, we measure regret, a standard metric in bandit problems that quantifies the difference between the cumulative reward an agent could have obtained by always selecting the optimal arm and the actual cumulative reward it accumulates over time.  Figure \ref{fig:bandit_results} presents the regret curves for this evaluation.

Following this, we conduct a finer-grained evaluation using two-armed Bernoulli bandits, introducing suboptimality gap analysis. The suboptimality gap $\Delta$ is defined as
\begin{equation}
\Delta := \theta^* - \theta,
\end{equation}
where $\theta^*$ is the success probability of the optimal arm, and $\theta$ is the success probability of the next-best arm. We evaluate the best-performing LLM strategy across three suboptimality gaps: $ \Delta \in \{0.2, 0.4, 0.6\} $. The $ \Delta = 0.2 $ condition presents a particularly challenging scenario, requiring models to distinguish between nearly identical reward probabilities. Figure \ref{fig:suboptimality} presents the regret curves for this evaluation.

\paragraph{Results.}
Our evaluation (Figure \ref{fig:bandit_results}) shows that explicit prompting significantly improves exploration efficiency, with GPT-4 achieving the lowest regret. Implicitly prompted models often commit to suboptimal arms early, suggesting they do not infer the need for exploration without direct instruction. From the suboptimality gap experiments (Figure \ref{fig:suboptimality}), we find that a prompted GPT-4 performs competitively with classical baselines when reward differences are distinct ($\Delta \ge 0.4$). However, it struggles in the more challenging $\Delta = 0.2$ setting, suggesting that while LLMs can be prompted to explore, their ability to distinguish between subtle statistical differences remains a limitation.

\begin{figure}[t]
    \centering
    \includegraphics[width=.85\linewidth]{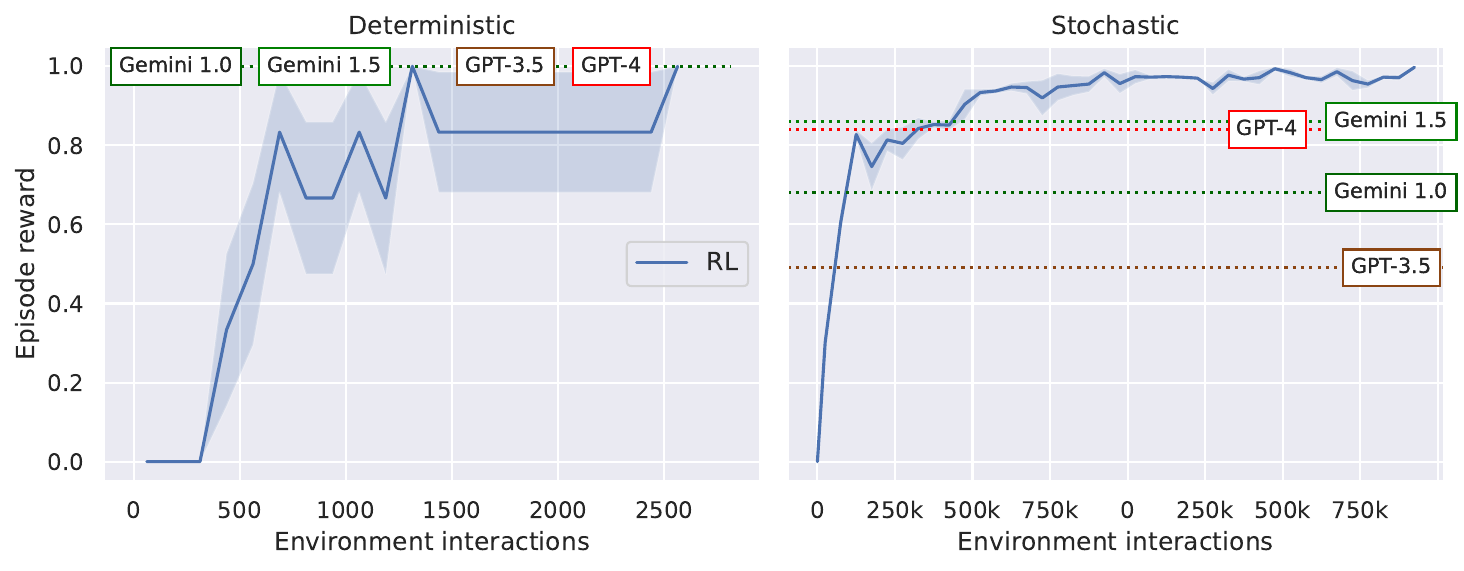}
    \includegraphics[width=.85\linewidth]{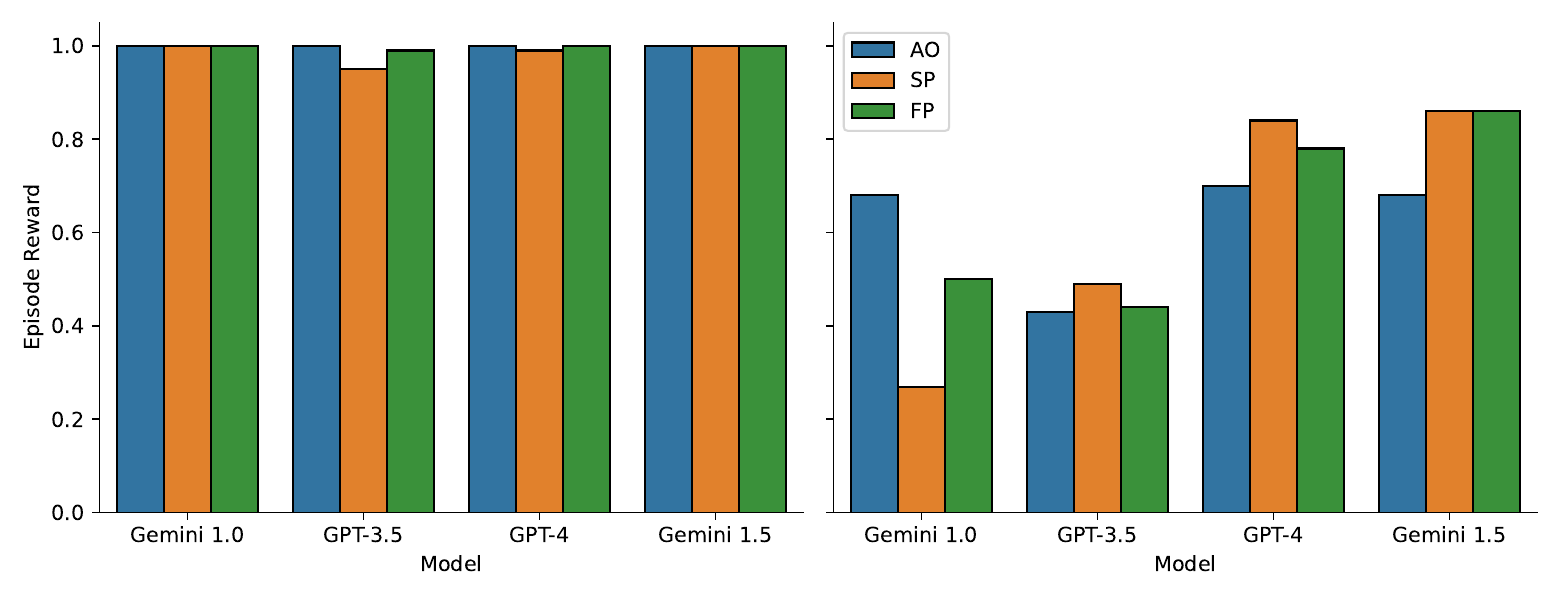}
    \caption{Decision-making results for the deterministic (left) and stochastic (right) setting. In the top row, the learning curves are visible for the RL agents. For each LLM, the performance of the best prompt approach is reported as a horizontal bar. In the bottom row, the performance for each prompt-model approach can be found for both settings.}
    \label{fig:decision-making-curves}
\end{figure}

\begin{figure}[h]
    \centering
    \includegraphics[width=0.85\linewidth]{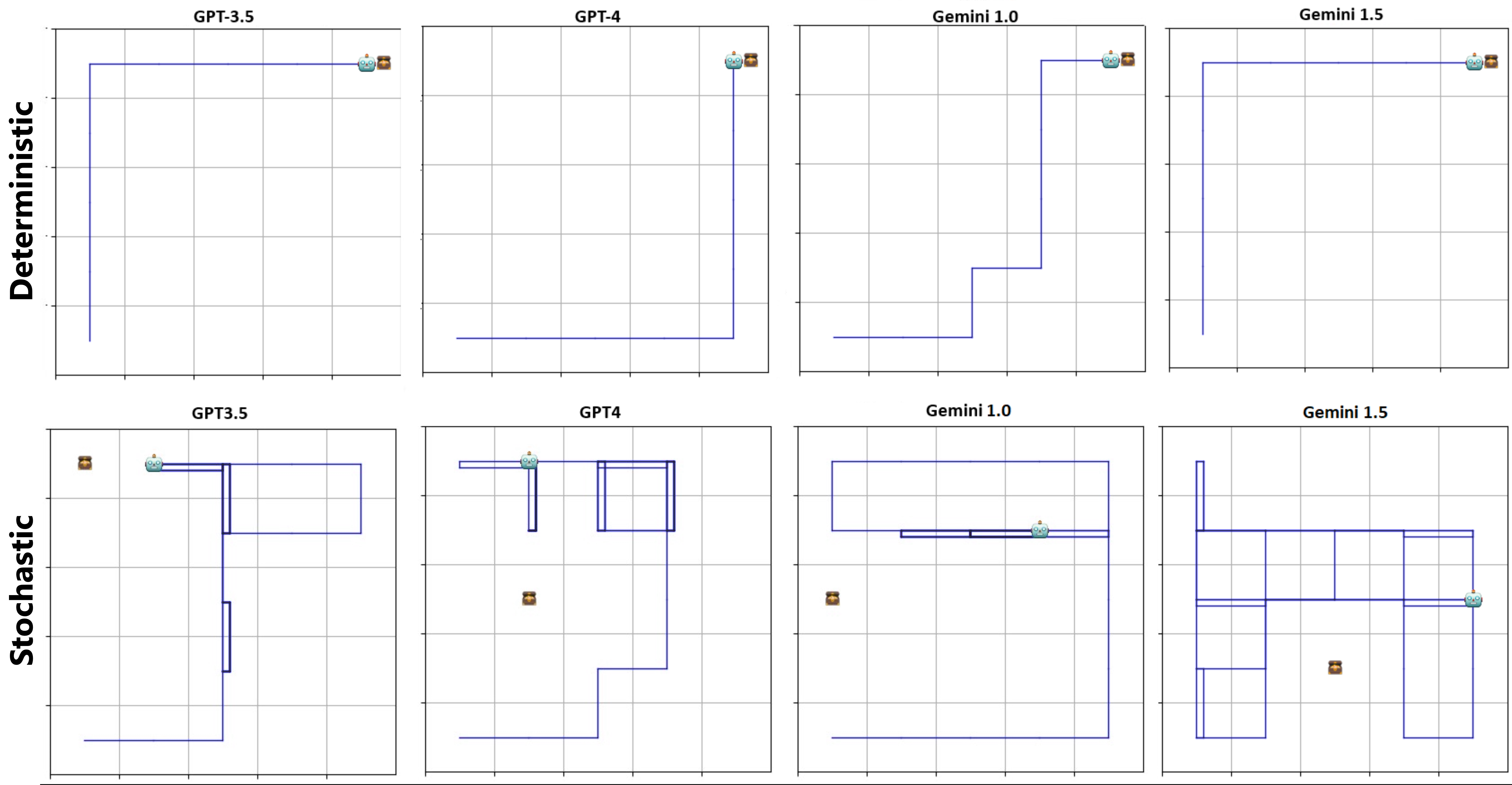}
    \caption{Example trajectories of various foundation model decision-making agents in the deterministic setting (top row) and the stochastic setting (bottom row).}
    \label{fig:trajectories_det}
\end{figure}

\subsection{Gridworld Environments} \label{sec:gridworld_experiments} To further evaluate how LLMs handle structured decision-making, we extend our analysis to Gridworld environments, where agents must explore a spatially structured state space instead of selecting from static actions. Unlike MABs, Gridworlds introduce state transitions, meaning optimal exploration requires both short-term action planning and long-term memory. This setting allows us to assess whether LLMs can reason about spatial navigation and exploration without direct supervision.

\paragraph{Gridworld Setup.}
We evaluate LLMs in two variations of a 5 × 5 Gridworld. First, in the deterministic Gridworld, the reward is placed in a fixed location, and the agent observes both its own coordinates and the reward’s location. This tests whether LLMs can efficiently navigate to a goal when full information is available. Then, in the stochastic Gridworld, the reward location is uniformly sampled at the start of each episode and is not visible to the agent. The agent only observes its own coordinates, making the problem partially observable and requiring systematic exploration to find the goal.

\paragraph{Prompts.}
To test whether LLMs can generalize across these settings, we compare three prompting strategies of increasing specificity. First, in Action Only (AO), the prompt provides minimal guidance, simply asking the agent to choose the next action. Second, in Simple Plan (SP), the agent is encouraged to reason about the next steps before selecting an action. Finally, Focused Plan (FP) explicitly instructs the agent to use the available memory of past visited locations to determine unexplored areas and plan the next movement. The complete prompt templates can be found in Appendix \ref{app:gridworld-decision-making}.

\paragraph{Algorithms.}
We compare the LLM agents to reinforcement learning baselines, using the Proximal Policy Optimization (PPO) algorithm in the deterministic setting and the RecurrentPPO adaptation in the stochastic setting \citep{schulman2017proximal}. RecurrentPPO leverages a recurrent neural network to incorporate temporal information, enabling it to handle partial observability effectively. We use the same setup of LLMs as the MABs experiments and average the performance across five random seeds.

\paragraph{Results.} Figure \ref{fig:decision-making-curves} presents the performance of each LLM-prompt strategy across both deterministic and stochastic settings, and in Figure \ref{fig:trajectories_det}, example trajectories of each of the foundation models can be found in both the deterministic and stochastic settings. In the deterministic setting, LLMs perform well across all prompting conditions, successfully navigating to the fixed reward location with minimal interactions. However, in the stochastic setting, where systematic exploration is required, performance declines sharply, particularly when using general prompts (Action-Only). Without explicit guidance, LLMs frequently revisit previously explored locations and fail to search the full state space efficiently. While the SP and FP prompts improve performance, they do not fully resolve the challenges LLMs face in handling long-horizon dependencies. Even with explicit instructions, LLMs currently struggle to effectively leverage memory over multiple interactions, leading to redundant exploration patterns. The RL baseline adapts to partial observability over time and eventually solves the environment. These experiments highlight the limitations of applying LLMs to structured exploration tasks requiring long-term memory.

\section{Zero-Shot VLM Performance in Atari} \label{sec:vlmatari}

\paragraph{Experimental Setup.} 
We evaluate GPT-4o on seven hard-exploration Atari games where traditional RL agents struggle due to sparse rewards: Freeway, Gravitar, Montezuma’s Revenge, Pitfall, Private Eye, Solaris, and Venture. To assess whether VLMs can infer objectives directly from visual input, we use a general, minimal prompt that remains the same across all games (Listing \ref{list:atari_prompt}), providing only the list of available actions. To contextualize performance, we compare its zero-shot cumulative reward against scores from a highly optimized RainbowDQN agent \citep{castro2018dopamine} at various stages of training (Table \ref{table:scores_vlm}).

\paragraph{Temporal Information.}
In RL, agents typically process stacks of consecutive frames to detect motion. However, VLMs must infer motion from visual cues alone. To accommodate this, we modify the standard approach by introducing a lag of $m=6$ timesteps between each of the four frames in the input stack. This increases temporal diversity, making motion easier to interpret semantically. Unlike other work \citep{waytowich2024atarigptbenchmarkingmultimodallarge}, our approach balances temporal variation and frame continuity to better capture exploration-relevant motion patterns.
\begin{figure}
\begin{lstlisting}[breaklines=true, language=JavaScript, caption=The prompt templates used for the Atari games., label={list:atari_prompt}]
Given the following four frames, where the last frame is the current game state, to beat the game - what action would you take from the actions {actions}? 
\end{lstlisting}
\end{figure}

\begin{table}[t]
\caption{Cumulative reward of GPT-4o compared to Rainbow (RB) at different stages of training and human scores across several hard-exploration environments.}
\label{table:scores_vlm}
\vskip 0.15in
\centering
{
\footnotesize
\begin{tabular}{lccccc}
\toprule
\textbf{Game} & \textbf{GPT-4o} & \textbf{RB 250K} & \textbf{RB 2.5M} & \textbf{RB 25M} & \textbf{Human} \\
\midrule    
Freeway  & 21 & 8 & 32 & 32 & {29.6}  \\
Gravitar & 500 & 64 & 199 & 2405 & {3351}   \\
Montezuma & 0 & 0 & 50 & 544 & {4753}  \\
Pitfall & -158 & -26 & -7 & -7 & {6464}  \\
Private Eye & -1000 & 503 & 125 & 1573 & {69571}   \\
Solaris & 600 & 681 & 1137  & 2093 & {12326} \\
Venture & 0 & 8 & 20 & 1513 & {1188}   \\
\bottomrule
 \end{tabular}
}
\vskip -0.1in
\end{table}

\begin{figure*}[t]
    \centering
    \includegraphics[width=0.49\linewidth]{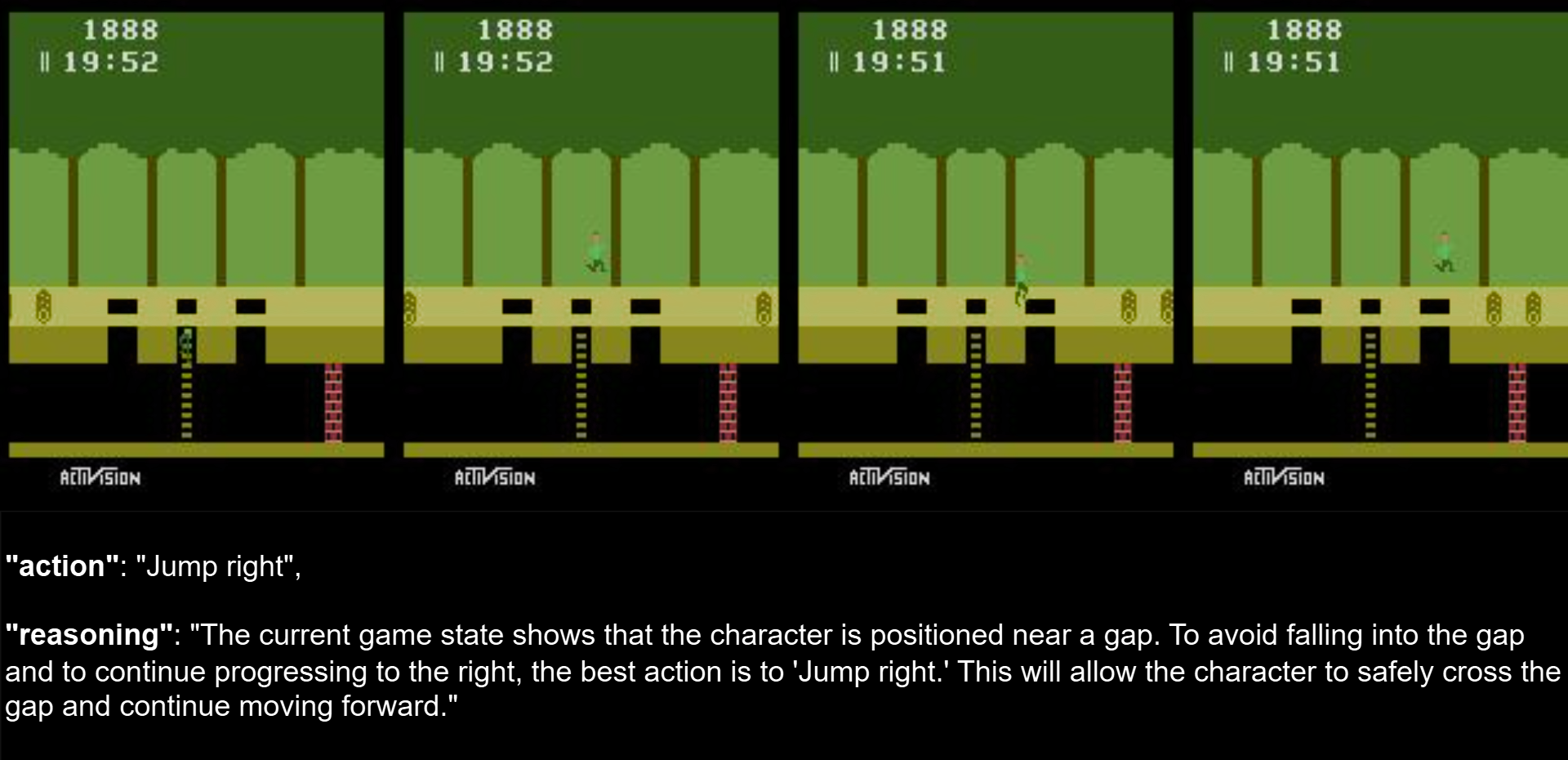}
    \includegraphics[width=0.49\linewidth]{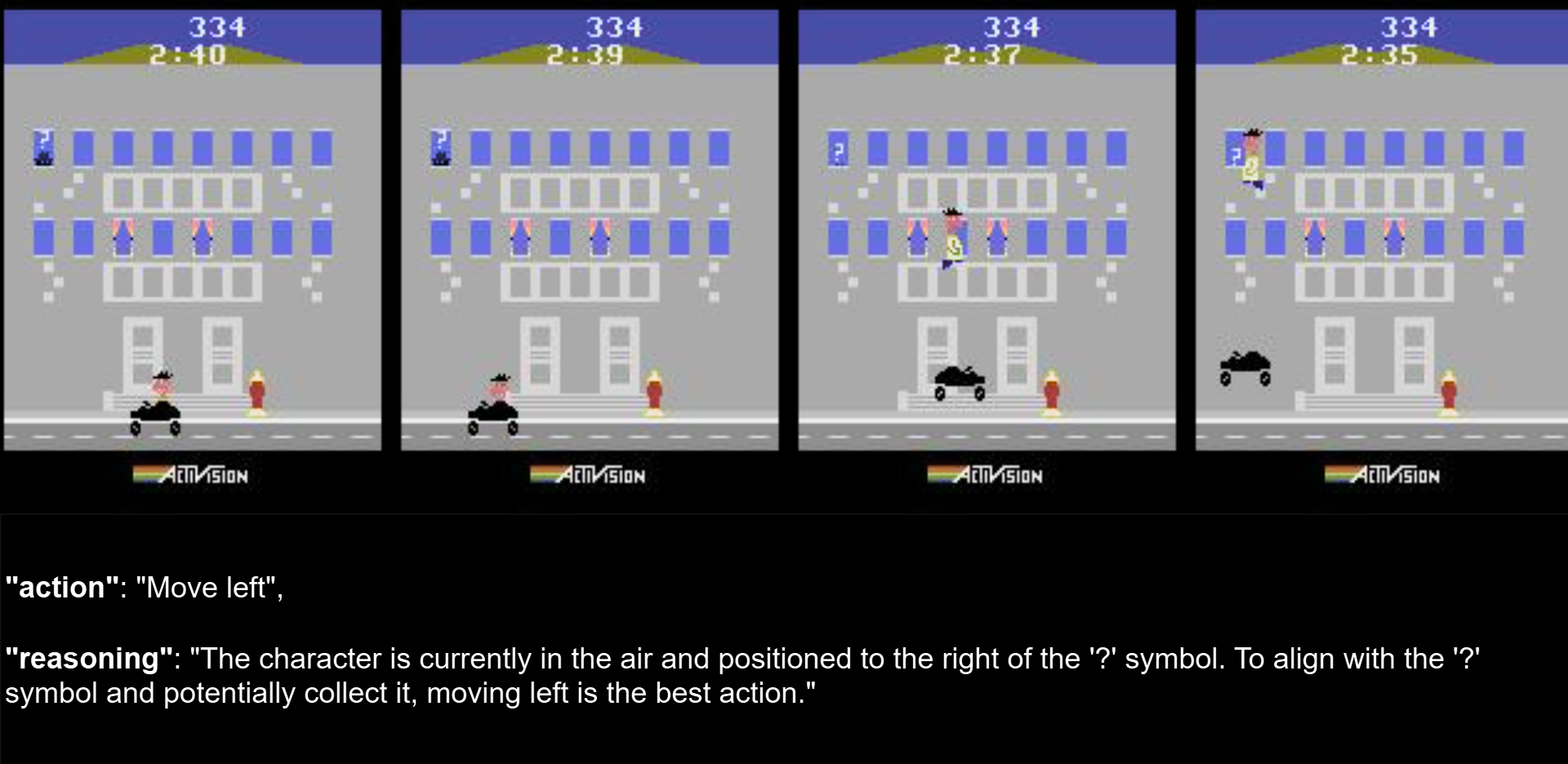}
    \includegraphics[width=0.49\linewidth]{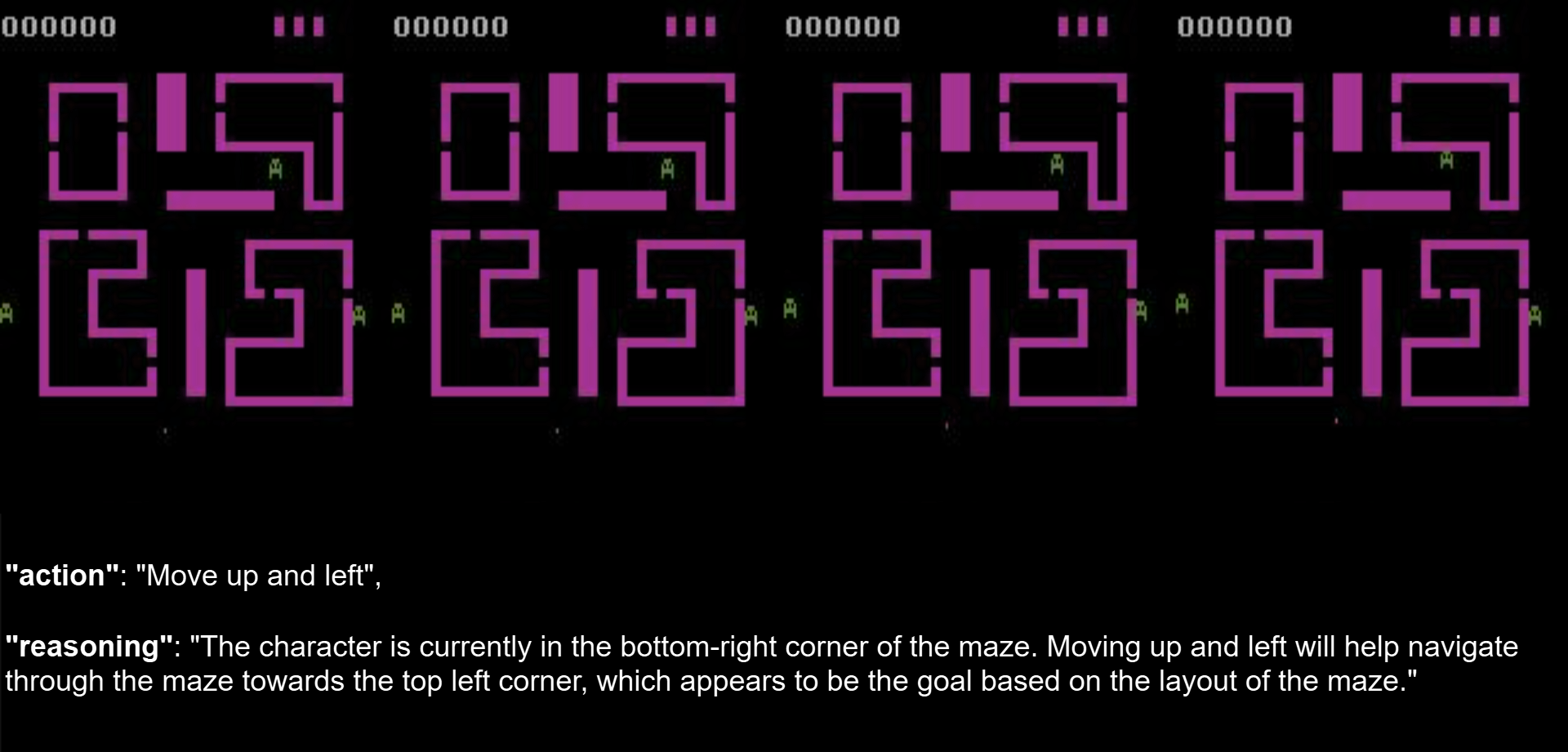}
    \includegraphics[width=0.49\linewidth]{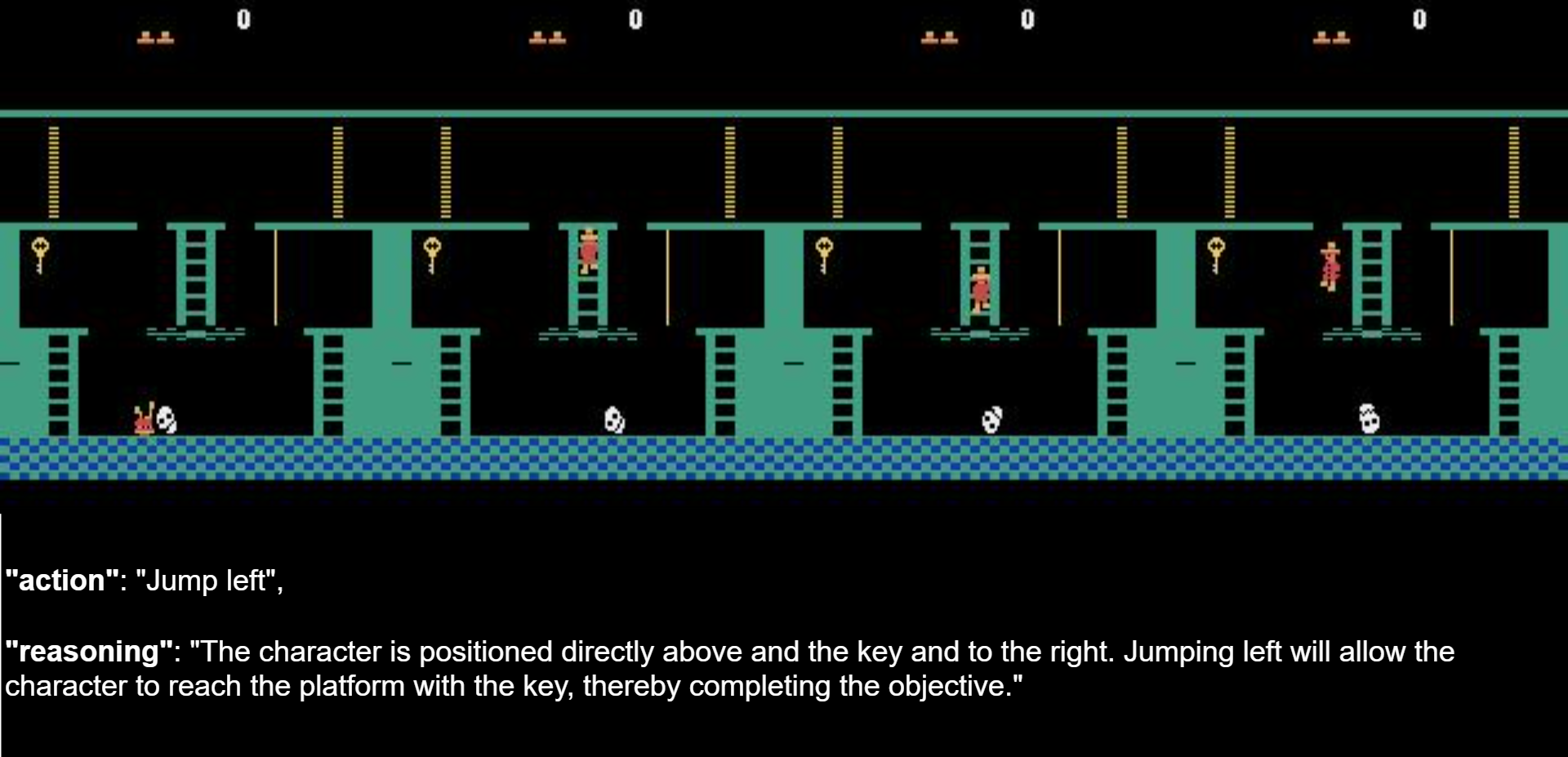}
    \includegraphics[width=0.49\linewidth]{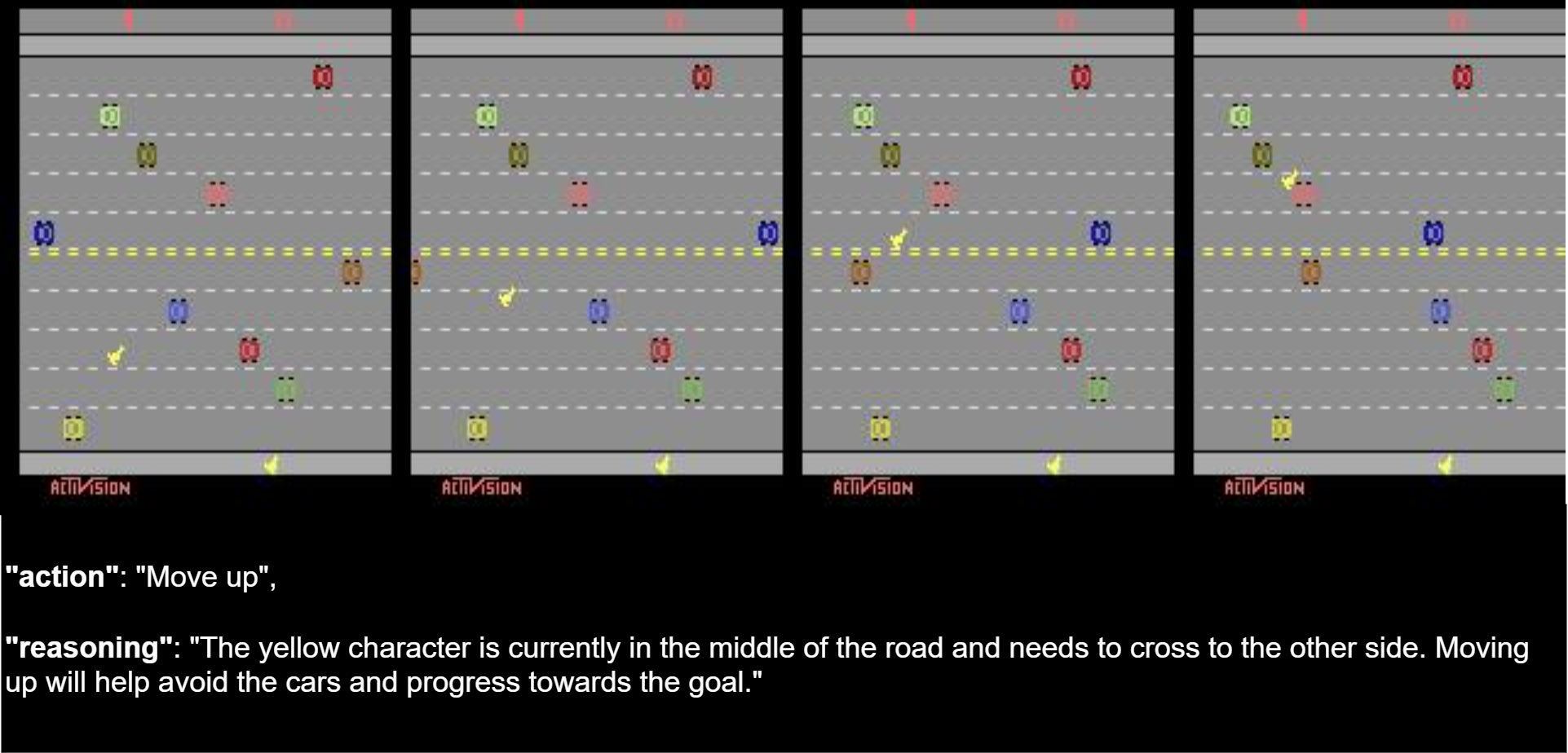}
    \includegraphics[width=0.49\linewidth]{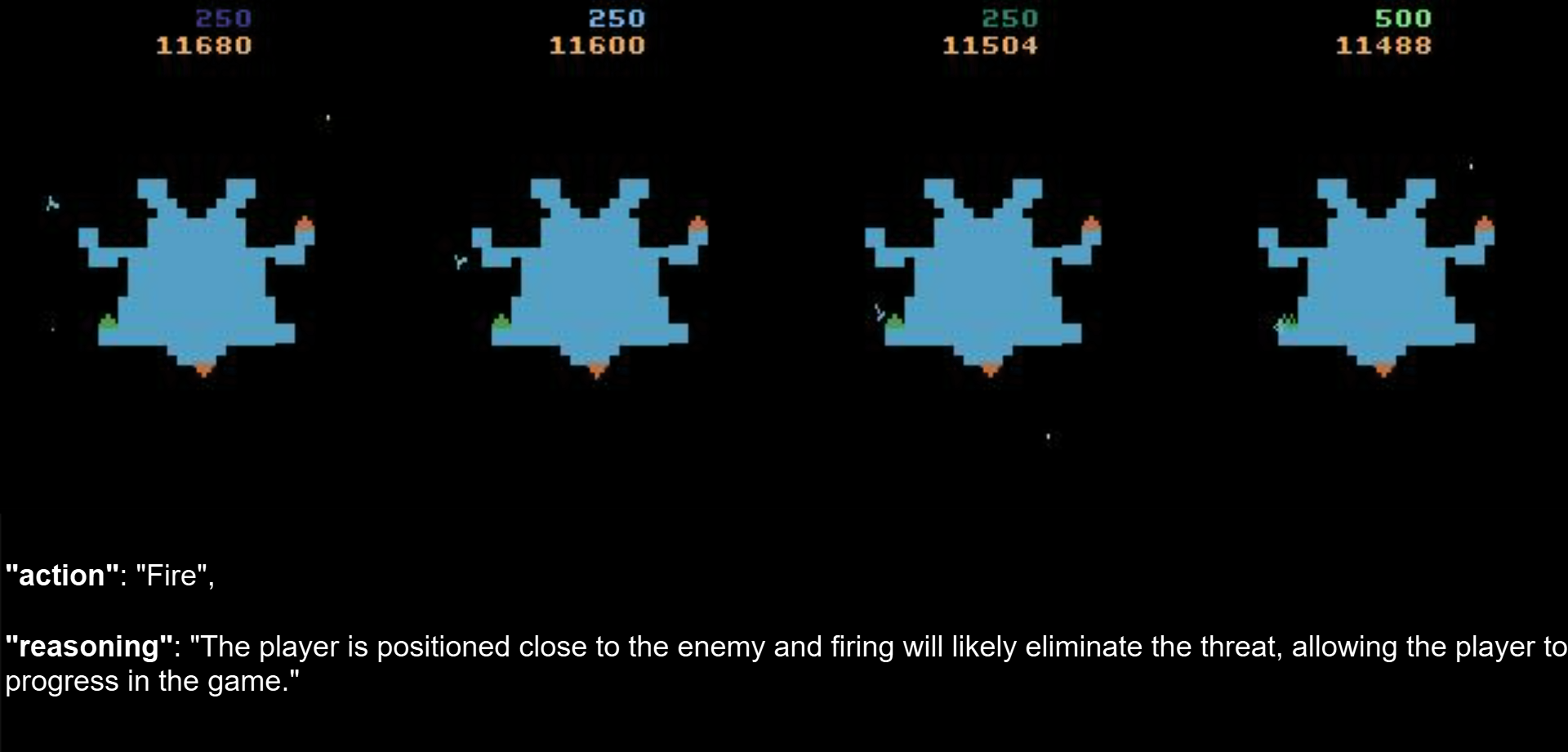}
    \includegraphics[width=0.49\linewidth]{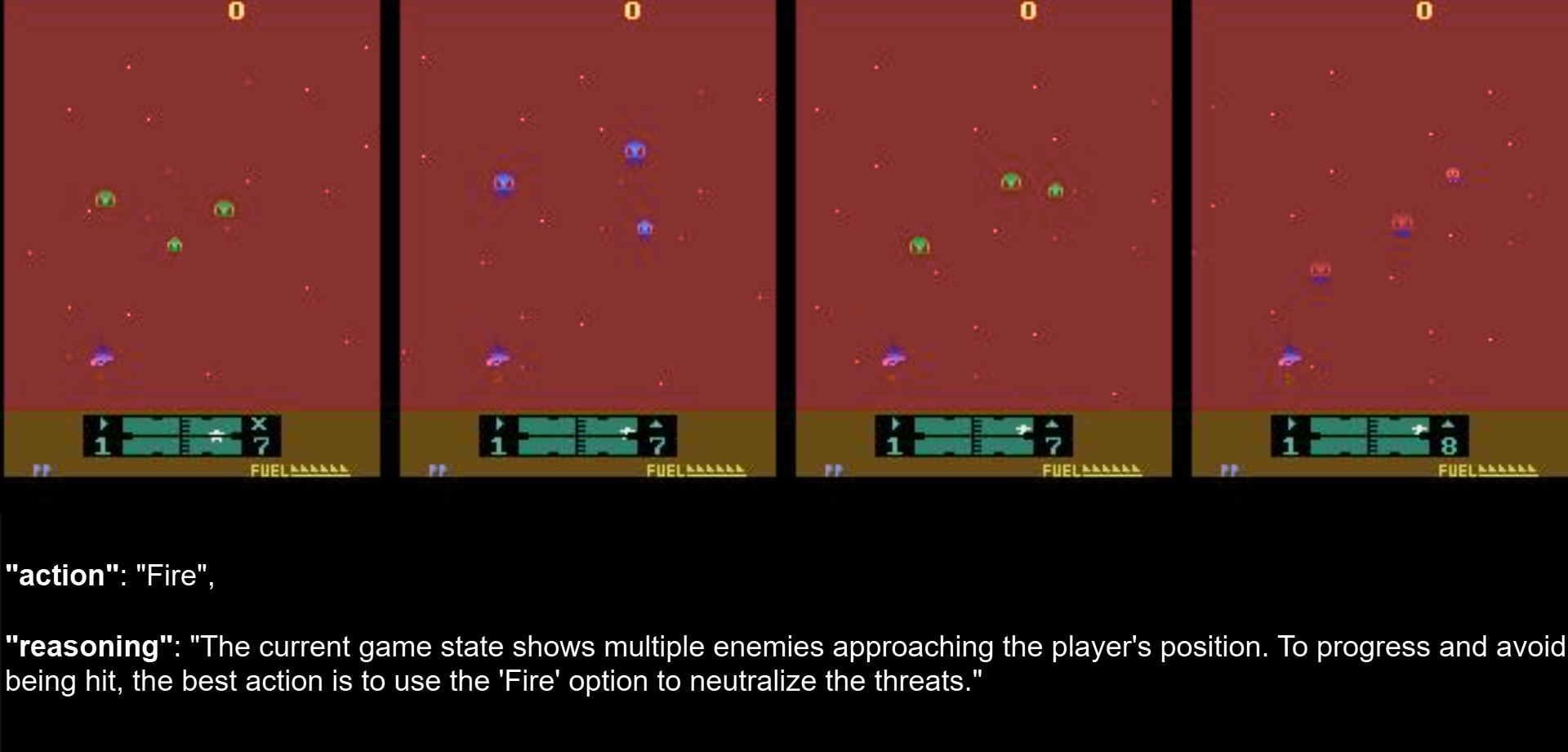}
    \caption{Selected moments from  all tested Atari games demonstrating both impressive high-level reasoning and poor low-level execution from the GPT-4o agent.}
    \label{fig:qualitative_vlm}
\end{figure*}
\paragraph{Quantitative Results.}
The quantitative results are presented in Table~\ref{table:scores_vlm}, comparing the zero-shot VLM agent (GPT-4o) against RainbowDQN at different stages of training, as well as human performance. We find that the VLM agent achieves strong zero-shot exploration in Freeway, Gravitar, and Solaris, obtaining cumulative rewards comparable to or exceeding those of Rainbow agents trained for hundreds of thousands of environment steps. In Freeway, the VLM agent quickly recognizes that moving up the road is the correct strategy, reaching 21 points—substantially outperforming Rainbow at 250K environment steps and approaching human-level performance. Similarly, in Gravitar, the VLM agent successfully navigates multiple levels, engages with enemy objects, and accurately fires its weapon, achieving 500 points, significantly surpassing Rainbow at 250K and 2.5M steps. In Solaris, the VLM agent performs comparably to Rainbow at 250K steps, demonstrating its ability to generalize without prior training.

\paragraph{Qualitative Analysis}

The divergence in quantitative performance motivates a deeper qualitative analysis to understand the underlying causes of VLM behavior. Our investigation, summarized in Figure \ref{fig:qualitative_vlm}, reveals a persistent "knowing-doing gap."

In several games, the VLM demonstrates an impressive ability to infer goals and strategies directly from pixels. In \textbf{Freeway}, it successfully recognizes the yellow character and understands its objective is to cross the road by moving up. The VLM's performance in \textbf{Gravitar} is particularly striking: it identifies the player's ship, its positioning relative to an enemy, and correctly deduces that it should fire, resulting in an immediate 250-point score. Similarly, in \textbf{Solaris}, the agent recognizes hostile ships and understands that the 'fire' action is required, even when enemies are not perfectly aligned in its path. These successes highlight a core strength: a powerful, pre-trained semantic understanding of objects and objectives.

This high-level understanding breaks down when precise, low-level control is required. In games like \textbf{Montezuma's Revenge} and \textbf{Pitfall}, the VLM correctly identifies the goal (e.g., "retrieve the key," "jump over the pit") but consistently fails at execution. It struggles with the precise timing and momentum needed to perform the actions it reasons about, leading to repeated failure. The gap is also evident in grounding and self-recognition. For example, in \textbf{Venture}, the VLM fails to identify the player's avatar (the small pink square), undermining any potential for strategic action.

Collectively, these successes and failures paint a clear picture. While VLMs possess a powerful semantic "knowing" of what to do, they often lack the fine-grained procedural "doing" required for execution. This provides strong motivation for investigating hybrid approaches that leverage their semantic guidance while offloading precise control to more robust learning mechanisms.

\section{An Upper-Bound Analysis of a Hybrid VLM-RL Agent}
\label{sec:hybrid_upper_bound}

Our analysis in Section \ref{sec:vlmatari} established that VLMs consistently fail as autonomous agents due to the knowing-doing gap. This motivates investigating hybrid frameworks—not as general solutions, but as tools to understand the theoretical upper bounds of VLM-RL synergies under ideal conditions.
.

\paragraph{Hybrid Algorithm.}
We propose a simple on-policy intervention where an RL agent's trajectory is periodically guided by a VLM, governed by an intervention probability $\epsilon$ and a duration $T$. We chose Proximal Policy Optimization (PPO) \citep{schulman2017proximal} as the base algorithm to explicitly test this on-policy intervention. The goal is not to distill the VLM's knowledge into a replay buffer, which would be a natural approach for an off-policy method. Instead, we use the VLM as a semantic explorer to steer the agent to a new state. The PPO agent then resumes its standard on-policy learning process from this more promising starting point.

\paragraph{Experimental Design.} To create a clean proof of concept, we selected \textit{Freeway}—an environment where the VLM's high-level strategy is known to be correct and the required control is simple. This allows us to cleanly isolate the potential effect of VLM guidance. We compare three agents:(1) a vanilla PPO baseline, (2) a PPO agent augmented with Random Network Distillation \citep{burda2019rnd} (PPO+RND) as a strong exploration baseline, and (3) our PPO-VLM hybrid. All agents were trained for 100,000 environment steps, with results averaged over 5 random seeds.

\begin{figure}[t]
    \centering
    \includegraphics[width=0.9\linewidth]{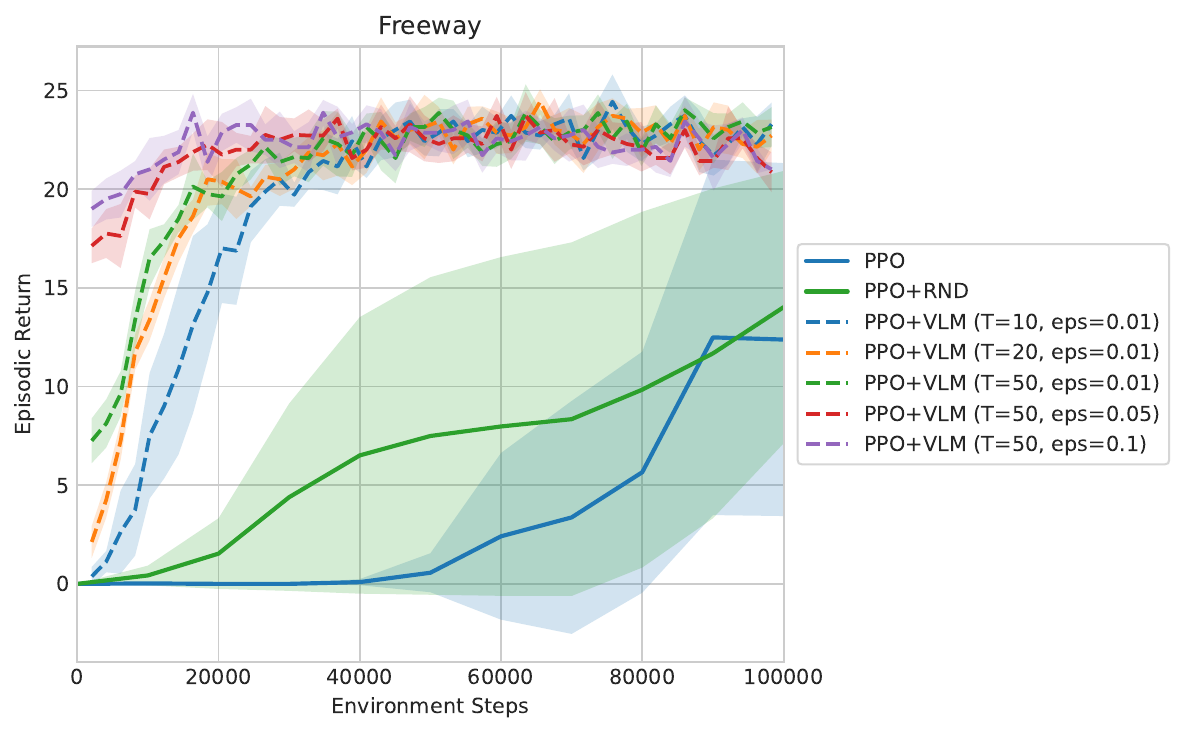}
    \caption{Training curves for various configurations of T and $\epsilon$ in Freeway}
    \label{fig:vlme}
\end{figure}

\paragraph{Analysis and Discussion.} The learning curves from our experiment are presented in Figure \ref{fig:vlme}. The data shows that in this specific setting, the PPO-VLM agent learns significantly faster than both the vanilla PPO and the strong PPO+RND baselines. This comes at the cost of increased computation per step due to VLM queries, creating a trade-off between sample efficiency and computational cost. Furthermore, this result should not be interpreted as a general solution, but as an empirical data point. It suggests that under favorable conditions, VLM guidance can act as a potent "semantic accelerator" for an RL policy. In more complex environments where a VLM's guidance is less reliable, we would expect a substantially smaller benefit. This study's contribution is to provide a clear, quantitative data point demonstrating that a synergy is possible under ideal conditions. Future research could extend this hybrid strategy to more complex settings where VLM guidance is less reliable, perhaps by developing adaptive scheduling mechanisms that integrate VLM assistance based on exploration uncertainty or by using policy distillation to mitigate the high inference cost of VLM queries.

\section{Conclusion}
\label{sec:conclusion}
Our systematic benchmark of foundation models in classic, hard-exploration RL tasks provides a clear characterization of their current capabilities and limitations. In text-based environments, we find that performance remains highly dependent on explicit instruction. In visually complex Atari games, our analysis reveals a persistent "knowing-doing gap," where VLMs consistently fail at the low-level execution and semantic grounding necessary for autonomous control, even when their high-level understanding appears correct.

Our investigation into a simple hybrid framework offers an encouraging data point. The results in Freeway serve as a potential upper bound on the sample efficiency gains, demonstrating that VLM guidance can significantly accelerate learning under favorable conditions. These findings indicate that while foundation models may not be ready to serve as end-to-end agents in complex domains, a promising path forward lies in designing hybrid systems for scenarios where VLMs demonstrate sound high-level reasoning but lack precise execution capabilities. Such systems can strategically leverage the semantic priors of foundation models to guide and bootstrap the learning of more robust, traditional RL policies, a direction that warrants further exploration.
\newpage

% \section*{Acknowledgments}
% Use unnumbered first level headings for the acknowledgments. All

% \section*{Ethics Statement}
% Authors can add an optional ethics statement to the paper. 
% For papers that touch on ethical issues, this section will be evaluated as part of the review process. The ethics statement should come at the end of the paper. It does not count toward the page limit, but should not be more than 1 page. 

% This tells LaTeX HOW to format the references
\bibliographystyle{plainnat}

% This tells LaTeX WHERE to find the reference entries.
% Make sure your .bib file is named "bib.bib" or change it here.
\bibliography{bib}

\newpage
\appendix
\section{Multi-Armed Bandit Experiments} \label{app:multiarmedbandits}
\subsection{Background}
\paragraph{Reinforcement Learning.} Reinforcement learning (RL) is a framework for sequential decision-making where an agent learns to maximize cumulative rewards by interacting with an environment. RL problems are typically modeled as a Markov Decision Process (MDP), defined by the tuple:
\begin{equation}
    (\mathcal{S}, \mathcal{A}, P, R, \gamma),
\end{equation}
where:
\begin{itemize}
    \item \( \mathcal{S} \) is the set of possible states of the environment.
    \item \( \mathcal{A} \) is the set of actions available to the agent.
    \item \( P(s' \mid s, a) \) is the transition probability function, defining the probability of moving to state \( s' \) given current state \( s \) and action \( a \).
    \item \( R(s, a) \) is the reward function, mapping state-action pairs to scalar rewards.
    \item \( \gamma \in [0,1] \) is the discount factor, determining the importance of future rewards.
\end{itemize}

At each timestep \( t \), the agent observes a state \( s_t \), selects an action \( a_t \), and transitions to a new state \( s_{t+1} \) based on the environment dynamics \( P \), receiving a reward \( R(s_t, a_t) \). The objective is to learn a policy \( \pi(a \mid s) \) that maximizes the expected return:
\begin{equation}
    G_t = \sum_{k=0}^{\infty} \gamma^k R(s_{t+k}, a_{t+k}).
\end{equation}

\paragraph{Multi-armed bandits.} The multi-armed bandit (MAB) problem is a fundamental decision-making framework in reinforcement learning. It models a scenario where an agent selects from a set of \( K \) independent actions, or ``arms,'' each associated with an unknown reward distribution. The goal is to maximize cumulative reward over a given horizon by balancing:
\begin{itemize}
    \item \textbf{Exploration}: Trying different arms to gather information about their reward distributions.
    \item \textbf{Exploitation}: Selecting the arm with the highest expected reward based on current knowledge.
\end{itemize}
Formally, at each time step \( t \), the agent selects an arm \( k \in \{1, ..., K\} \), receiving a reward \( r_t \) drawn from an unknown distribution \( P_k \):
\begin{equation}
    r_t \sim P_k.
\end{equation}
In our experiments, we specifically consider \textit{Bernoulli bandits}, where each arm provides rewards sampled from a Bernoulli distribution with an unknown success probability \( \theta_k \). That is, for each arm \( k \),
\begin{equation}
    r_t \sim \text{Bernoulli}(\theta_k),
\end{equation}
where \( \theta_k \) represents the probability of obtaining a reward of 1, while a reward of 0 occurs with probability \( 1 - \theta_k \).

The agent's objective is to identify the optimal arm \( k^* \) with the highest \( \theta_k \), while minimizing cumulative regret over time. The suboptimality gap \( \Delta_k \) for an arm \( k \) is defined as:
\begin{equation}
    \Delta_k = \theta^* - \theta_k,
\end{equation}
where \( \theta^* = \max_k \theta_k \) is the success probability of the optimal arm.

\subsection{Experiments}
\paragraph{Algorithms.} For the multi-armed bandit experiments, methods like Thompson Sampling and UCB naturally consider previous trials' outcomes by updating their belief distributions. As could be seen in Listing \ref{list:bandit_prompt}, to account for this in the decision-making for the LLMs, we provide a memory in the prompt. In this memory, we append all previous trials as '\textit{Pulled arm \{ACTION\} resulting in a reward of \{REWARD\}}'. In the case of Thompson Sampling, we found that the best performing prior was $\alpha=1$ and $\beta=1$, and in the case of UCB, we used the UCB1 variant with a tuned constant $c = 0.25$.

\paragraph{Prompt phrasing.} In preliminary experiments, we tried slight variations of the prompt presented in Listing \ref{list:bandit_prompt}. For instance, we found that providing the maximum number of trials or encouraging "an efficient exploration approach" did not improve the performance. 

\begin{figure}[h]
    \centering
    \includegraphics[width=\linewidth]{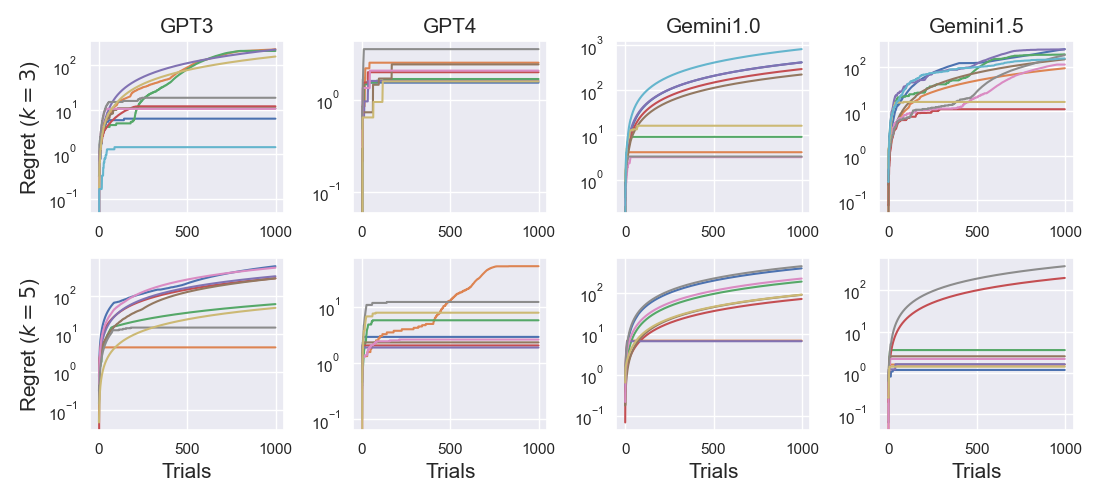}
    \caption{Individual seeds for each of the LLMs using the best-performing prompt in the $k=3$ (top) and $k=5$ (bottom) settings, providing insights on the different strategies employed by each model.}
    \label{fig:seeds}
\end{figure}

\paragraph{Additional Results.} Additional results can be found in Figure \ref{fig:seeds} where the individual seeds for the best-performing LLM prompts can be found in the two bandit settings.

\section{Gridworld Decision-Making Experiments} \label{app:gridworld-decision-making}
\paragraph{Reinforcement learning agents.} \label{app:strategies} For the reinforcement learning agents in these experiments, we used the default Stable-Baselines3 implementations of PPO and RecurrentPPO \citep{stable-baselines3}.

\paragraph{Prompt templates.} 
For the foundation models, the prompt templates used for the deterministic and stochastic LLM agents are almost identical. While the deterministic agent receives the exact position where the reward is located, the stochastic agent is only told the following: \textit{Your goal is to reach the reward located at a random coordinate as quickly as possible.} See Listing \ref{listing_llm_agent_prompt} for the full prompt used by the deterministic LLM agent. The agent's memory is filled as it interacts with the environment. Whenever an action is executed, we add the following line to the memory: "\textit{Executed \{ACTION\} at \{LOCATION\} resulting in \{NEW LOCATION\} and no reward.}" Additionally, as seen in Listing \ref{listing_simple_plan}, whenever an agent chooses an action, it outputs a plan representing its thoughts. We add each plan to the memory as well.

\begin{figure}[H]
\begin{lstlisting}[breaklines=true, language=JavaScript, caption=The prompt template for the deterministic LLM agent., label={listing_llm_agent_prompt}]
**Context**
You are an agent in a {n}x{n} grid.
The bottom left corner is at {BOTTOM LEFT}, top left at [0, n-1], top right at [{n-1, n-1}], and bottom right at [{n-1}, 0].
The x-axis increases as you move rightward, and the y axis increases as you move upwards. 
Your goal is to reach the reward located at a coordinate [{n-1},{n-1}] (the top-right corner).

**Memory**
[...]

**Observation**
Your current location is {OBSERVATION}

**Available Actions**
up
right
down
left

**Task**
Choose an action from the given list of actions. Output your response using the following JSOn format and do not use markdown.
{OUTPUT FORMAT}
\end{lstlisting}
\end{figure}

The three prompting approaches are implemented using different JSON output formats. Note that since the action-only agent outputs no plan, its memory will only contain the results of the executed actions. Note that in the stochastic setting, we do not mention that the reward is located in the top-right corner.

\begin{figure}[H]
\begin{lstlisting}[breaklines=true, language=JavaScript, caption=The output format used by the \textbf{Action Only} agent.]
{
    "action": "{The action you want to take}"
}
\end{lstlisting}
\end{figure}

\begin{figure}[H]
\begin{lstlisting}[breaklines=true, language=JavaScript, caption=The output format used by the \textbf{Simple Plan} agent., label={listing_simple_plan}]
{
    "plan": "{Think about what you want to do next to fulfill your goal.}",
    "action": "{The action you want to take}"
}
\end{lstlisting}
\end{figure}

\begin{figure}[H]
\begin{lstlisting}[breaklines=true, language=JavaScript, caption=The output format used by the \textbf{Focused Plan} agent.]
{
    "plan": "{Use your memory to determine which position you want to go next.}",
    "analysis": "{Using your plan analyze which action is best to efficiently reach the position.}",
    "action": "{The action you want to take}"
}
\end{lstlisting}
\end{figure}

\paragraph{Results.}
The full list of numerical results for the FA performances from the experiments in Section \ref{sec:gridworld_experiments} can be found in Table \ref{tab:agent_performances}. 

\begin{table}[h]
    \caption{LLM Agent performances for an empty $5\times5$ grid with a fixed reward location and random reward location, averaged over 100 episodes.}
    \vspace{0.1in}
    \centering
        \begin{tabular}{lcc}
        \toprule
        \textbf{Model} & \textbf{Fixed Reward} & \textbf{Random Reward} \\
        \midrule
        Gemini 1.0 (AO) & 100\%  & 68\% \\
        Gemini 1.0 (SP) & 100\% & 27\%\\
        Gemini 1.0 (FP) & 100\% & 50\% \\
                \hline
                
        GPT-3.5 (AO) & 100\% & 43\% \\
        GPT-3.5 (SP) & 95\% & 49\%  \\
        GPT-3.5 (FP) & 99\% &  44\%\\
                \hline

        GPT-4 (AO) & 100\% & 70\% \\
        GPT-4 (SP) & 99\% & 84\% \\
        GPT-4 (FP) & 100\% & 78\% \\
                \hline

        Gemini 1.5 (AO) & 100\% & 68\% \\
        Gemini 1.5 (SP) & 100\% & 86\% \\
        Gemini 1.5 (FP) & 100\% & 86\% \\
        \bottomrule
        \end{tabular}
    \label{tab:agent_performances}
\end{table}

\section{Atari Experiments}

\paragraph{Algorithms.} For the Atari experiments, we passed four individual frames with a lag of 6 environment timesteps between using a frameskip of 4. Along with these frames, we provided the prompt and action space to the OpenAI API. The action space was provided as specified by the official ALE action space descriptions. The RainbowDQN implementation and results are from the Dopamine \citep{castro2018dopamine}, while the PPO implementation came from the default visual implementation from Stable-Baselines3 \citep{stable-baselines3}. For the hybrid strategy we implemented the VLM actions in the Stable-Baselines3 implementation in an epsilon-greedy fashion.

\section{Compute} \label{app:compute}
We use OpenAI’s APIs for GPT-3.5, GPT-4, and GPT-4o, and Google Studio's APIs for Gemini 1.0 and Gemini 1.5. 
For training the PPO, we used a single NVIDIA A100 GPU in all environments. For the GPT models, we used 'GPT-3.5-turbo-0613', 'GPT-4-0613', and 'GPT-4o-2024-08-06' cutoffs, which cost US\$ 0,50 / 1M input tokens US\$ 1,50 / 1M output tokens, US\$ 30,00 / 1M input tokens US\$ 60,00 / 1M output tokens, US\$ 2,50 / 1M input tokens US\$ 10,00 / 1M output tokens, respectively, as of writing. 
The Gemini models can be used freely as of writing, although the Gemini models have a relatively low limit for queries per minute. For the Gemini models we used 'gemini-1.0-pro-001' and 'gemini-1.5-pro'.

\end{document}